\documentclass{article}

\usepackage{amsmath,amsfonts,amssymb}
\usepackage{wrapfig}
\usepackage{graphicx}
\usepackage{booktabs}
\usepackage{enumitem} 
\usepackage{subcaption} 

\usepackage{threeparttable} 

\usepackage[final]{corl_2025} 

\title{ Learning Impact-Rich Rotational Maneuvers via Centroidal Velocity Rewards and Sim-to-Real Techniques: A One-Leg Hopper Flip Case Study
}

\author{Dongyun Kang\thanks{These authors contributed equally.}$^{*1}$,
Gijeong Kim$^{*1}$,
JongHun Choe$^{*1}$,
Hajun Kim$^{1}$,
Hae-Won Park$^{1}$ \\
\\
$^{1}$Korea Advanced Institute of Science and Technology (KAIST)
}

\begin{document}
\maketitle

\begin{figure}[!h]
  \centering
  \includegraphics[width=\columnwidth]{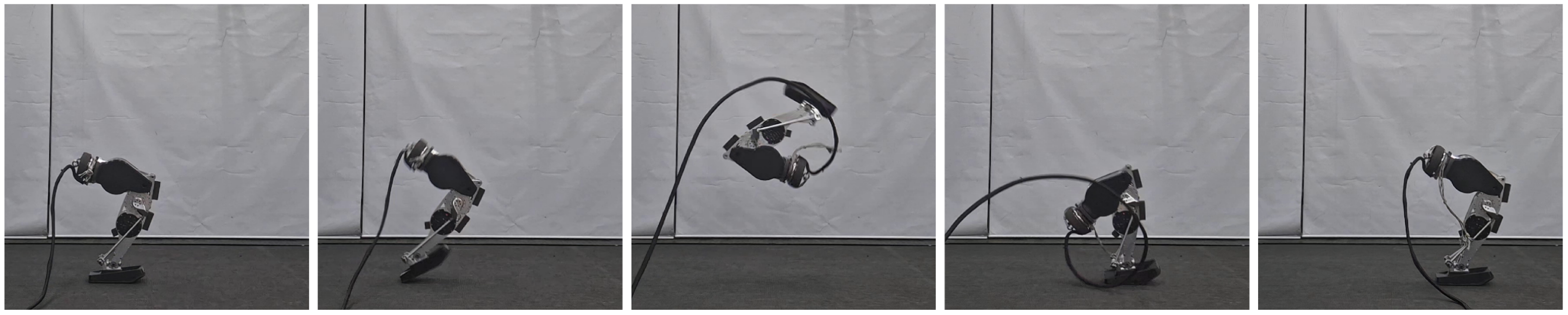}
  \caption{Snapshots of the first successful deployment of the learned front flip on the real one-leg hopper platform. The policy was trained with a centroidal velocity--based reward and sim-to-real techniques to achieve robust, impact-rich rotation on hardware.}
  \label{fig:motion snapshot}
\end{figure}

\begin{abstract}
Dynamic rotational maneuvers, such as front flips, inherently involve large angular momentum generation and intense impact forces, presenting major challenges for reinforcement learning and sim-to-real transfer. In this work, we propose a general framework for learning and deploying impact-rich, rotation-intensive behaviors through centroidal velocity-based rewards and actuator-aware sim-to-real techniques. We identify that conventional link-level reward formulations fail to induce true whole-body rotation and introduce a centroidal angular velocity reward that accurately captures system-wide rotational dynamics. To bridge the sim-to-real gap under extreme conditions, we model motor operating regions (MOR) and apply transmission load regularization to ensure realistic torque commands and mechanical robustness. Using the one-leg hopper front flip as a representative case study, we demonstrate the first successful hardware realization of a full front flip. Our results highlight that incorporating centroidal dynamics and actuator constraints is critical for reliably executing highly dynamic motions.
A supplementary video is available at: \url{https://youtu.be/atMAVI4s1RY}
\end{abstract}



\section{Introduction}
\label{sec:introduction}
Legged robots have recently demonstrated an impressive range of dynamic behaviors, from agile locomotion to acrobatic maneuvers~\cite{jin2022high,MOR2024,hoeller2024anymal,atlas2021doubleflip,unitree2025sideflip,atlas2025sideflip,kim2024stage,Cafempc2024}. These capabilities push beyond low-speed locomotion and begin to exploit the full dynamics of the system.
Critically, such dynamic behaviors inherently involve the generation and dissipation of momentum through strong impacts~\cite{kim2024stage,Cafempc2024}, the modulation of body inertia through posture changes~\cite{atlas2021doubleflip,unitree2025sideflip}, and the precise control of rotational motion during flight or contact transitions~\cite{atlas2025sideflip}. Managing these phenomena—momentum exchange and inertial shaping—is a hallmark of dynamic legged systems, distinguishing them from other robotic platforms~\cite{ha2024learning}.
However, robotic motions that demand large body rotations and endure strong impacts pose fundamental challenges for reinforcement learning (RL) and sim-to-real transfer. Policies must not only coordinate whole-body dynamics to achieve significant rotational maneuvers~\cite{BD_centroidal,MIThumanoid2021}, but must also respect tight hardware limits on torque production~\cite{MIThumanoid2021,MOR2024}, and impact tolerance~\cite{ostyn2024improving}. Small mismatches between simulation and reality can easily result in catastrophic failures during such high-energy motions.

In this work, we propose a framework for learning and deploying impact-rich, rotation-intensive behaviors through centroidal velocity–based rewards and actuator-aware sim-to-real techniques.
We identify that conventional link-level reward formulations often fail to induce true whole-body rotation, as they can be satisfied by exploiting internal joint motions without achieving meaningful global movement. To address this, we introduce a centroidal angular velocity reward that accurately captures system-wide rotational dynamics and drives the discovery of flipping behaviors.
To bridge the sim-to-real gap under extreme actuation demands, we incorporate Motor Operating Region (MOR) modeling into the simulation. By accurately reflecting the torque-speed limitations of real actuators, this ensures that policies remain within feasible motor capabilities throughout training. To protect vulnerable hardware components during high-impact contacts, we apply transmission load regularization, penalizing excessive joint stresses and promoting safer landing strategies.

As a concrete demonstration, we select the front flip on a one-leg hopper, a minimal legged system with a single foot, a small support polygon, and no rotational authority from auxiliary limbs.
Performing a front flip on such a platform is challenging due to the highly nonlinear coupling between leg actuation and body rotation, the need for precise impulse generation during takeoff, and the demanding contact dynamics at landing.
We validate our framework through the first successful hardware realization of a front flip on this minimal system, demonstrating that impact-rich, rotation-intensive behaviors can be learned and deployed reliably under severe actuation and structural constraints.
This task epitomizes the challenges of executing large aerial rotations and absorbing high-magnitude impacts, making it an ideal benchmark for evaluating our methodology.
Moreover, mastering this maneuver provides broader insights into controlling underactuated robotic systems, offering a transferable recipe for robust sim-to-real learning beyond the specific case of flipping.

In summary, our contributions are threefold:
\begin{itemize}[leftmargin=*, itemsep=0pt, topsep=0pt]
\item We propose a centroidal angular velocity reward that accurately captures whole-body rotation and guides the discovery of dynamic maneuvers.
\item We introduce sim-to-real transfer techniques, including realistic Motor Operating Region (MOR) modeling and transmission load regularization, and analyze their critical role in achieving realistic, and hardware-resilient deployment in impact-rich rotational tasks.
\item We demonstrate the first successful hardware implementation of a front flip on a one-leg hopper, validating the effectiveness of our approach on a minimal and challenging platform.
\end{itemize}

\section{Related Work}
\label{sec:relatedwork}




\textbf{Dynamic Motion in Legged Robots.}
In the legged robot fields, many studies have explored dynamic motions such as fast running~\cite{park2017high,jin2022high,MOR2024}, bucking~\cite{kim2025contact}, parkour~\cite{caluwaerts2023barkour,zhuang2023robot,hoeller2024anymal}. In particular, some studies have demonstrated agile rotational maneuvers, such as cartwheels~\cite{atlas2025sideflip}, barrel rolls~\cite{atlas2021doubleflip}, and flips~\cite{unitree2025sideflip,atlas2017backflip,kim2024stage,li2023learning,fuchioka2023opt,Cafempc2024,mit2019backflip}. These challenging maneuvers involve high impacts and require consideration of complex dynamics during both takeoff and landing.

\textbf{Motion Planning with Centroidal Momemtum.} 
The centroidal momentum model has been extensively studied in model-based motion generation approaches~\cite{RMP2007,CMM2008,BD_centroidal,MIThumanoid2021,momentumAwareTO2022}, as it effectively captures the whole-body dynamics of legged robots~\cite{wensing2023optimization}.  
Orin et al.~\cite{CMM2008} analyzed the Centroidal Momentum Matrix (CMM) and characterized the concept of centroidal velocity.  
Dai et al.~\cite{BD_centroidal} and Zhou et al.~\cite{MIThumanoid2021} proposed motion planning frameworks that incorporate centroidal momentum for acrobatic motion in humanoid robots.
In learning-based methods, base angular velocity rewards have been widely used to encourage rotational motion~\cite{kim2024stage,vezzi2024two}.  
More recently, several studies have begun to utilize centroidal momentum to better capture whole-body dynamics~\cite{ha2024learning,ferigo2021emergence,zhang2024achieving,xie2025humanoid}.  
In particular, Xie et al.~\cite{xie2025humanoid} introduced rewards based on centroidal angular momentum to account for the contribution of leg-induced momentum.  
However, these approaches have primarily focused on momentum regularization for balance control, rather than on actively generating dynamic rotational behaviors. In contrast, our work leverages the concept of centroidal velocity to explicitly guide policy learning toward rotation-intensive motions.

\textbf{Sim-to-Real Transfer in Dynamic and Impact-Rich Tasks.}
Applying learned motions to real legged robots remains highly challenging due to the sim-to-real gap~\cite{ha2024learning}. 
To address this, various methods have been proposed to mitigate this gap, including domain randomization~\cite{tan2018sim,bin2020learning,lee2020learning,miki2022learning}, noise injection~\cite{lee2020learning,miki2022learning,relaxedLogBarrier2024}, and online system identification using real-world data~\cite{chen2024identifying,kim2025online}. As the field progresses, and legged robots advance toward more dynamic behaviors, recent studies increasingly focus on handling high impacts and actuator constraints.
For instance, \citet{MOR2024} adopted MOR constraints to consider the actuator's physical limits during training, while \citet{ostyn2024improving} proposed an impact-aware planner that accounts for transmission-level load propagation. Building on these efforts, in this work, we investigate sim-to-real transfer techniques that explicitly incorporate actuator constraints, aiming to enable reliable execution of dynamic and impact-intensive motions on real legged robots.

\section{Hardware Platform}
\label{sec:hardware}
\begin{wrapfigure}{r}{0.16\textwidth}     
\vspace{-20pt}
\centering     
\includegraphics[width=0.16\textwidth]{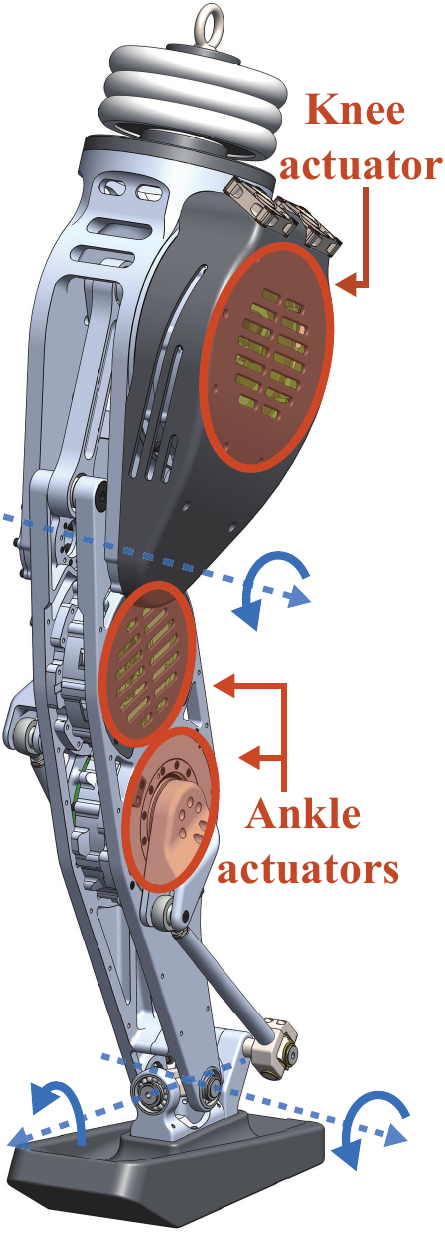} 
\caption{One-leg hopper used in this study.}
\vspace{-20pt}
\end{wrapfigure}

Our target platform is a compact, custom‐designed 3-DOF one-leg hopper system inspired by human lower‐limb kinematics.
It consists of a single knee pitch joint and a two‐axis ankle (pitch and roll), terminating in a flat foot.
The platform is capable of executing forward, backward, and lateral command-tracking locomotion, and can recover from external pushes with no additional support.
Key specifications and structural features are as follows:

\textbf{Physical Dimensions.}
The robot weighs 12.45 kg, and when its knee is fully extended, it reaches 890 mm in height. The thigh is 0.35 m long and weighs 7.60 kg, the calf is 0.41 m long and weighs 3.89 kg, and the foot is 0.08 m long in height and weighs 0.96 kg. These measurements closely match human lower-limb proportions. The flat foot measures approximately 200 mm in length and 100 mm in width. And a small toe section is integrated into the foot plate to increase contact area and improve ground reaction stability.

\textbf{Closed-Loop Ankle Mechanism.}
Ankle pitch and roll motions are produced by two parallel actuators coupled through a closed-loop mechanism using universal and ball joints. 
This enforces kinematic constraints between the actuators and the foot. The closed-loop structure is simulated in RaiSim~\cite{hwangbo2018per}, which provides pin constraints in the kinematic tree to enforce coupling consistency.


\section{Method}
\label{sec:method}
\subsection{Centroidal Velocity based Reward} \label{subsec:centroidalvelocityreward}
Maximizing base angular velocity (BAV) is a common reward design for rotational motions such as flips~\cite{kim2024stage,vezzi2024two}.  
However, we found that such link-level rewards can fail to induce true whole-body rotation.  
In a one-leg hopper, rewarding the base (i.e., the thigh) pitch angular velocity leads the robot to generate rapid internal motion between the thigh and calf without net body rotation (see Sec.~\ref{subsec:CAV Result}).
Because the internal angular momenta of the thigh and calf cancel, the robot achieves high base angular velocity while producing negligible net angular momentum.  
An alternative is to maximize centroidal angular momentum (CAM).  
While CAM-based rewards encourage momentum generation, we observed that they fail to convert it into sufficient angular velocity for completing the flip.
This is because increasing rotation speed requires not only generating momentum but also reducing inertia, yet CAM rewards provide no incentive for inertia reduction. As a result, policies trained with CAM rewards converge to inefficient motions that do not leverage inertia modulation to amplify rotation speed.
In this work, we instead use a reward based on the centroidal angular velocity (CAV), which inherently accounts for both angular momentum and moment of inertia.

Centroidal dynamics describe the motion of a multibody system in terms of translation and rotation about its center of mass (COM). According to \cite{CMM2008}, the centroidal momentum $\mathbf{h_G} = [\mathbf{p_{com}}^T,\;\mathbf{L_{com}}^T]^T \in \mathbb{R}^6$ satisfies the relation $\mathbf{h_G} = \mathbf{I_G} \mathbf{v_G}$, where $\mathbf{v_G} = [\mathbf{v_{com}}^T,\;\mathbf{w_{com}}^T]^T \in \mathbb{R}^6$ is the system’s 6D twist, and $\mathbf{I_G} \in \mathbb{R}^{6 \times 6}$ is the centroidal composite rigid body inertia. $\mathbf{I_G}$ represents the total inertia of the system when all joints are locked, effectively modeling it as a rigid body.
When $\mathbf{I_G}$ is defined at the COM, it takes a block-diagonal form: $\mathbf{I_G} = \operatorname{blockdiag}(m \mathbf{I_3}, \mathbf{I_{com}})$, where $m$ is the total mass and $\mathbf{I_{com}} \in \mathbb{R}^{3 \times 3}$ is the composite rigid body inertia about the COM. The angular momentum component therefore reduces to $\mathbf{L_{com}} = \mathbf{I_{com}} \mathbf{w_{com}}$, and the centroidal angular velocity (CAV), which can be interpreted as the system’s overall rotation rate, is given by $\mathbf{w_{com}} = \mathbf{I_{com}^{-1}} \mathbf{L_{com}}\in\mathbb{R}^3$~\cite{CMM2008}.
We use both the centroidal angular momentum (CAM) $\mathbf{L_{com}}$ and the centroidal angular velocity (CAV) $\mathbf{w_{com}}$ to formulate our flip reward. The two-second motion is divided into three phases: a takeoff phase (0–0.5 seconds), an aerial phase (0.5–1.05 seconds), and a landing phase (1.05–2.0 seconds). The aerial phase duration can be adjusted for different tasks, allowing for longer airborne motions by extending flight time.
During the aerial phase, we encourage angular velocity aligned with the desired rotational axis $\boldsymbol{\alpha}$, and during landing, we penalize residual angular momentum. For front flips, we set $\boldsymbol{\alpha} = [0,\ 1,\ 0]^T$ to target pitch rotation. 
The reward is defined as follows: $r_{\mathrm{CAV}} = 0$ during takeoff, $r_{\mathrm{CAV}} = \max(\min(\boldsymbol{\alpha}^T \mathbf{w}_{\mathrm{com}}, 10), -0.1)$ during the aerial phase, and $r_{\mathrm{CAV}} = -0.5\, \min(\| \mathbf{L}_{\mathrm{com}} \|,\ 2.5)$ during landing.

\subsection{Sim-to-Real Transfer and Impact Mitigation Techniques}
\label{subsec:simtorealtechniques}
In this section we outline two key techniques that enable reliable transfer of the learned flip motion from simulation to the physical one-leg hopper. Additional techniques used to facilitate sim-to-real transfer are described in Appendix~\ref{app:auxiliary sim2real}.

\begin{wrapfigure}{r}{0.35\textwidth}     
\vspace{-20pt}
\centering     
\includegraphics[width=0.35\textwidth]{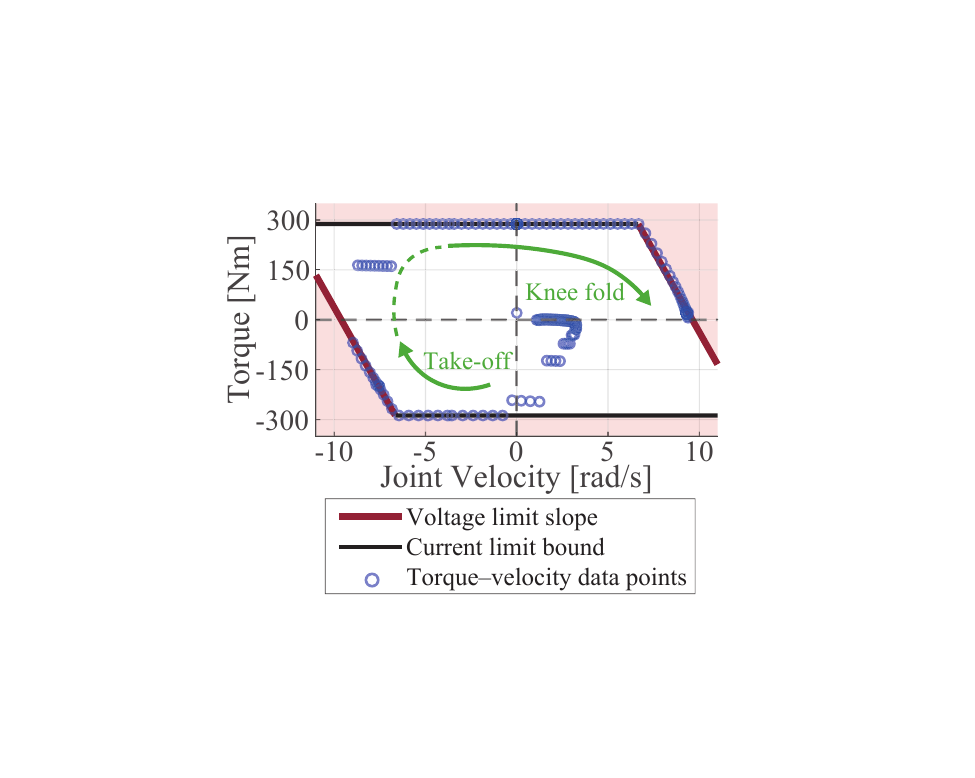} 
\caption{Motor Operating Region (MOR) at the knee actuator. Simulation data were collected within $\pm0.25$ seconds during the flip motion around take-off, with take-off and knee fold events also noted in the figure. Red regions denote areas beyond the MOR.}
\label{fig:mor_envelope}
\vspace{-20pt}
\end{wrapfigure}

\textbf{Motor Operating Region.}
Dynamic, impact‐rich maneuvers demand torque performance at the very edge of actuator capabilities. In particular, front flips on a one leg hopper require rapid takeoff impulses that push the knee and ankle motors into high‐speed, high‐torque regimes. Therefore, accurately modeling the available Motor Operating Region (MOR)~\cite{MOR2024,MIThumanoid2021} and constraining commands within this envelope during simulation is critical for reliable policy transfer.

Many reinforcement learning frameworks adopt a simple box‐type MOR, with constant torque and velocity bounds. Such fixed bounds fail to capture the speed‐dependent characteristics of real motors, where available torque drops with increasing rotor speed due to back‐EMF.
Fig.~\ref{fig:mor_envelope} illustrates a more realistic MOR~\cite{MOR2024}, shown as a trapezoidal region in the torque–speed plane. This envelope is bounded by two critical lines.  
The first boundary, marked in red, is the \textit{voltage limit slope}. As rotor speed increases, the back-EMF reduces the available voltage, causing a linear falloff in maximum torque. 
The second boundary, shown in black, is the \textit{soft current limit}. Even if the voltage margin allows greater torque, a current limit is enforced to prevent overheating.
During training, we clip commanded torques to remain within this envelope. Without this constraint, the policy exploits torque–speed combinations that are unattainable on the real hardware.  
The blue points in Fig.~\ref{fig:mor_envelope} represent torque–speed pairs commanded during the flip motion. Notably, the policy naturally drives the actuator near the MOR boundary, fully utilizing the available region to generate sufficient takeoff impulse.  
If a simplified box-type MOR were used instead, the policy would learn in an unrealistic region (red region in Fig.~\ref{fig:mor_envelope}). On real hardware, this mismatch may cause incorrect contact impulses at takeoff, leading to poor transfer and failed flips.

\textbf{Transmission Load Regularization.}
To protect vulnerable transmission components, we introduce a regularization term that penalizes large joint loads induced by contact impacts. The one-leg hopper experiences high-impact forces on a single limb, and early hardware tests revealed sun gear failure due to excessive impulse-induced loads.  Notably, minimizing contact impulses alone does not guarantee reduced actuator loads in the closed-loop ankle linkage, since impact forces can redistribute through the mechanism.
We estimate the instantaneous transmission load from external contacts using the contact impulse and jacobian. These external torques are then averaged over the control interval to compute the effective joint load:
$\boldsymbol{\tau}_{\mathrm{inst}} = \sum_{\text{contacts}} \frac{\mathbf{J}^\mathsf{T}\, \boldsymbol\lambda}{\Delta t_{\mathrm{sim}}}, \quad
\boldsymbol{\tau}_{\mathrm{load}} = \frac{1}{\Delta t_{\mathrm{control}}/\Delta t_{\mathrm{sim}}} \sum \bigl| \boldsymbol{\tau}_{\mathrm{inst}} \bigr|$
where \( \mathbf{J} \) is the contact Jacobian, \( \boldsymbol{\lambda} \) is the contact impulse, \( \Delta t_{\mathrm{sim}} \) is the simulation timestep, and \( \Delta t_{\mathrm{control}} \) is the control interval.  
The operator \( \lvert \cdot \rvert \) denotes component-wise absolute value.
We employ the relaxed log barrier function~\cite{relaxedLogBarrier2024}
to constrain the peak transmission load.
This regularization term is added to the barrier reward to guide the policy towards landing strategies that reduce extreme transmission stress.
Furthermore, we impose episode termination for extreme transmission overload. Although the barrier penalty discourages high loads, the brief duration of impact can lead policies to tolerate occasional violations in pursuit of long-term return. To reflect irreversible hardware damage, we terminate the episode with 50\% probability whenever \(\lvert\boldsymbol{\tau}_{\mathrm{load}}\rvert\) exceeds a critical threshold. This probabilistic termination preserves exploration during early training while imposing a realistic cost for severe overloads that would otherwise disable the real hardware.  

\subsection{Implementation Details}
\textbf{Reward Composition.}  
The total reward is composed of two parts: \emph{motion rewards}, which guide behavior discovery, and \emph{barrier rewards}, which enforce hardware constraints. The motion reward includes the centroidal angular velocity term (Sec.~\ref{subsec:centroidalvelocityreward}) and regularization penalties for stabilizing behavior. The barrier reward employs relaxed log barriers~\cite{relaxedLogBarrier2024} applied to joint limits and transmission load (Sec.~\ref{subsec:simtorealtechniques}). These are combined as scalar values via  
\( r_{\mathrm{motion}} = (1 + r_{\mathrm{CAV}})\sum_{i} w_i r_i \) and  \( r_{\mathrm{barrier}} = \sum_{j} w_j b_j \) where \(r_i\) and \(w_i\) denote the \(i\)th motion reward term and its weight, and \(b_j\) and \(w_j\) denote the \(j\)th barrier term and its weight.
Details of individual terms are listed in Appendix~\ref{app:reward formulation}.

\textbf{Policy Architecture.}  
The policy is trained using Proximal Policy Optimization (PPO) with an asymmetric actor–critic structure. The actor receives proprioceptive observations, short action history, and phase. The critic is given privileged information, including unactuated joint states and body linear velocity. To bridge the observation gap, we adopt a concurrent estimation framework~\cite{Concurrent2022} that augments the policy input with predicted privileged states. See Appendix~\ref{app:training} for further details.

\section{Results}
\label{sec:result}

\subsection{Effect of Centroidal Angular Velocity based Reward} \label{subsec:CAV Result}

\begin{figure}[h]
  \centering
  \includegraphics[width=1.0\columnwidth]{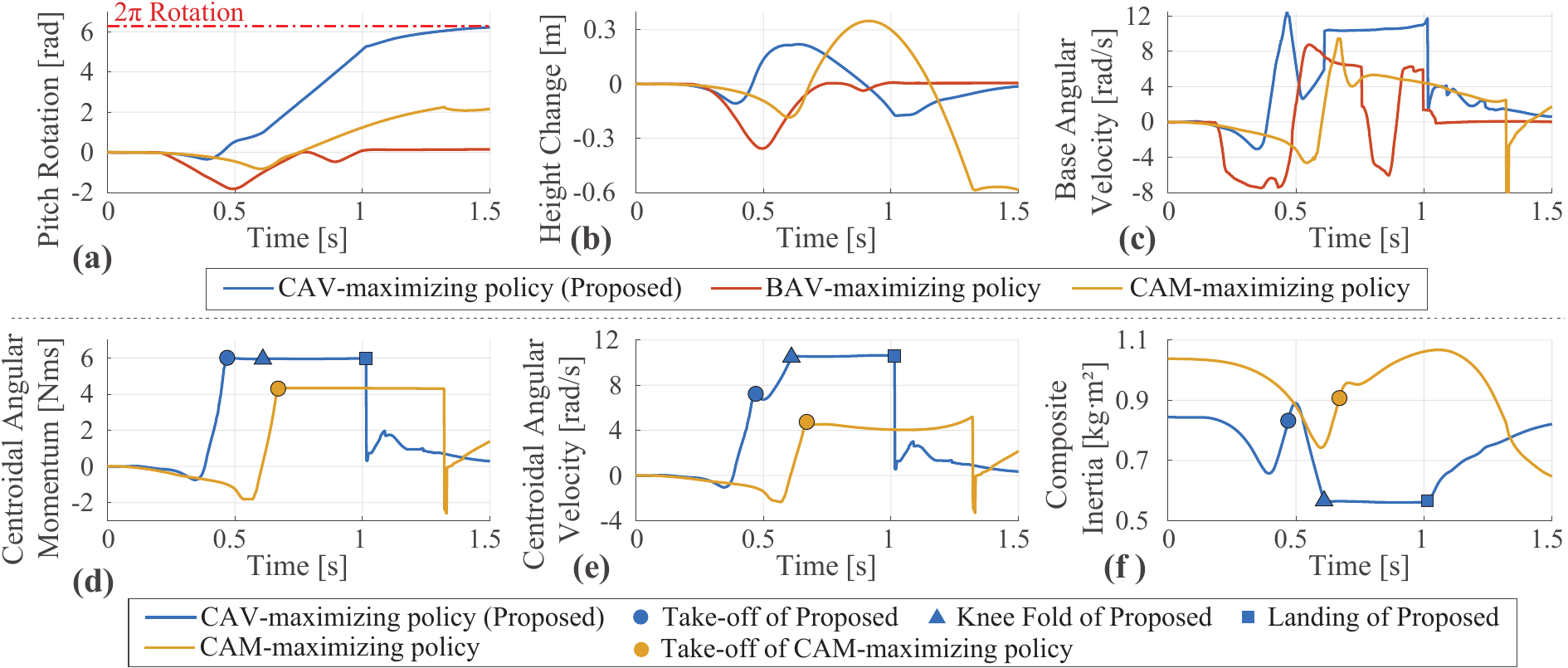}
  \caption{Comparison of policies trained with different aerial-phase rewards (0.5–1.05 s) for front-flip motion. (a)–(b) show base pitch rotation and vertical height (relative to initial pose); (c) shows base angular velocity about the pitch axis. (d)–(f) present centroidal angular momentum (CAM), centroidal angular velocity (CAV), and composite pitch-axis inertia for CAM- and CAV-maximizing policies. Key events (take-off, knee fold, landing) are marked in (d)–(f) for both policies.}
  \label{fig:reward_effects}
\end{figure}

To compare the effects of different reward formulations on rotational motion, we train policies to maximize base angular velocity (BAV), centroidal angular momentum (CAM), and centroidal angular velocity (CAV, proposed). Each reward replaces the aerial-phase term (0.5–1.05 seconds) in $r_{\mathrm{CAV}}$ to induce a front flip. Fig.~\ref{fig:reward_effects} illustrates the resulting motion for each policy.

The BAV-maximizing policy executes a rapid knee flexion–extension motion between a fully extended singular pose and a folded configuration. Base angular velocity is generated through knee extension from the folded posture, reaching 8.7~rad/s (Fig.~\ref{fig:reward_effects}c). Despite this high base rotational speed, the robot fails to achieve take-off (Fig.~\ref{fig:reward_effects}b) or produce net whole-body rotation (Fig.~\ref{fig:reward_effects}a). 
In contrast to the BAV-maximizing policy, the CAM-maximizing policy initiates take-off (Fig.~\ref{fig:reward_effects}b) and generates nonzero net rotation. However, the rotation is insufficient to complete a full flip (Fig.~\ref{fig:reward_effects}a). The key limitation arises from maintaining a large composite inertia through an extended leg configuration, to achieve high angular momentum despite low rotational speed. This allows the policy to reach a centroidal angular momentum of 4.3N$\cdot$s, which is comparable to that of the proposed policy (6.0N$\cdot$s; Fig.\ref{fig:reward_effects}d). However, the resulting centroidal angular velocity remains substantially lower (4.8rad/s vs.\ 10.6rad/s; Fig.\ref{fig:reward_effects}e). Since only the magnitude of angular momentum is rewarded, the policy does not attempt to accelerate its rotation mid-flight by reducing inertia. As a result, the composite inertia remains large throughout the aerial phase (Fig.~\ref{fig:reward_effects}f), and no post-take-off configuration change occurs that would increase rotational speed. The proposed CAV-maximizing policy completes a full front flip (Fig.\ref{fig:reward_effects}a). Prior to take-off, the robot extends the knee while applying force through the heel, generating substantial centroidal angular momentum (Fig.\ref{fig:reward_effects}d). Once airborne, this momentum is conserved in the absence of external torques. The policy then rapidly flexes the knee to reduce the composite rotational inertia (Fig.\ref{fig:reward_effects}f), resulting in a 43$\%$ increase in centroidal angular velocity, from 7.4rad/s at take-off to a peak of 10.6rad/s during flight (Fig.\ref{fig:reward_effects}e). This mid-air acceleration enables the robot to complete a full 360-degree rotation before landing (Fig.\ref{fig:reward_effects}a-b). These results indicate that optimizing centroidal angular velocity enables both momentum utilization and inertia modulation, which are critical for executing dynamic rotational maneuvers.

\subsection{One-Leg Hopper Front Flip Results and MOR} \label{subsec:MOR Result}

\begin{figure}[t]
  \centering
  \includegraphics[width=0.98\columnwidth]{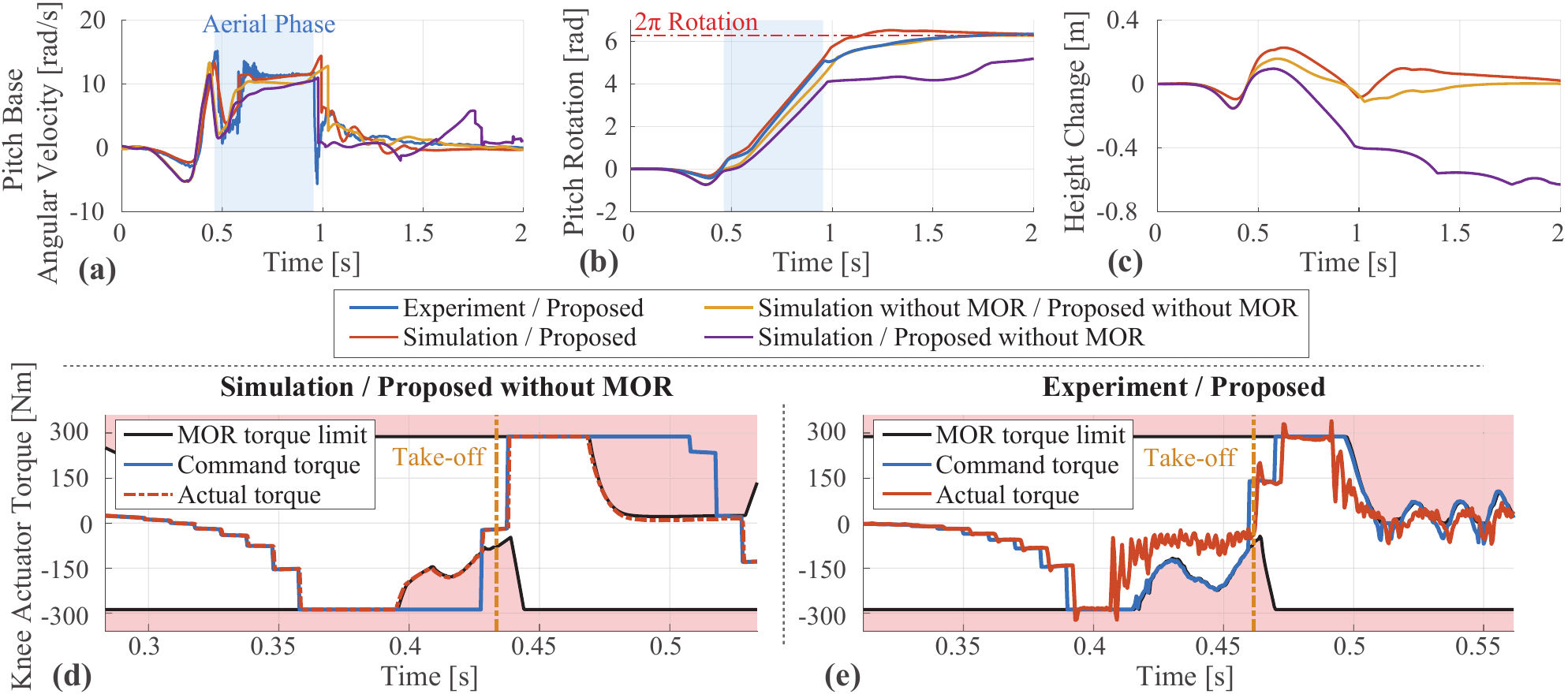}
  \caption{Comparison of simulation and hardware results for policies with and without MOR constraints. (a–c) show base pitch angular velocity, pitch rotation, and vertical height (simulation only), respectively. (d)–(e) show knee torque around take-off from policies trained without MOR (simulation) and with MOR (hardware experiment), respectively. Shaded regions indicate MOR violations.}
  \label{fig:experiments_data}
\vspace{-10pt}
\end{figure}

We deployed the proposed policy on the one-leg hopper hardware, achieving a full front flip (Fig.~\ref{fig:experiments_data}). To evaluate the role of MOR constraints in enabling such dynamic behavior, we compare four cases: The proposed policy, trained with MOR constraints, is evaluated in (1) hardware and (2) simulation with MOR constraints. A second policy, trained without MOR, is evaluated in simulation (3) without and (4) with MOR constraints. In the MOR-unconstrained setup, only static torque and velocity limits from the URDF are enforced, forming a box-shaped constraint in the torque–velocity space.

The policy trained without MOR constraints fails to complete a front flip when evaluated under MOR constraints, despite succeeding in the MOR-unconstrained simulation used for training. As shown in Fig.\ref{fig:experiments_data}a–c, the robot generates insufficient rotational velocity and momentum, resulting in landing before completing a full rotation. This failure arises from torque commands that exceed the feasible MOR, particularly during take-off and mid-air inertia modulation (Fig.\ref{fig:experiments_data}d). Along the MOR, the actuator’s available torque decreases at high joint velocities, when torque and velocity are aligned (i.e., during positive mechanical work). These limits are critical during take-off, when the policy attempts to extend the knee and push off with the ankle at high speed, and again after take-off when the knee is rapidly flexed to reduce composite inertia. In both phases, Fig.~\ref{fig:experiments_data}d show a reduction in allowable torque, which the policy, trained under static box-shaped torque limits, fails to respect. This mismatch produces unrealizable torque commands, resulting in reduced actual torque (Fig.~\ref{fig:experiments_data}d), insufficient rotational velocity (Fig.~\ref{fig:experiments_data}a), and ultimately failure to complete a full rotation (Fig.~\ref{fig:experiments_data}b). In contrast, the proposed method incorporates MOR constraints during training, allowing the policy to adapt to the reduced torque capacity near joint speed extremes observed during take-off and mid-air reconfiguration. As a result, in real-world experiments, the commanded torques remain within the actuator’s feasible envelope and yield measured torques that follow the intended profiles (Fig.\ref{fig:experiments_data}e). By generating executable torque commands, the proposed method reproduces the simulated angular velocity profile (Fig.\ref{fig:experiments_data}a) and completes a full front flip on hardware (Fig.\ref{fig:experiments_data}b).

\subsection{Effect of Transmission Load Regularization} \label{subsec:TLR Result}
\begin{wrapfigure}{r}{0.38\textwidth}
    \centering
    \includegraphics[width=0.36\textwidth]{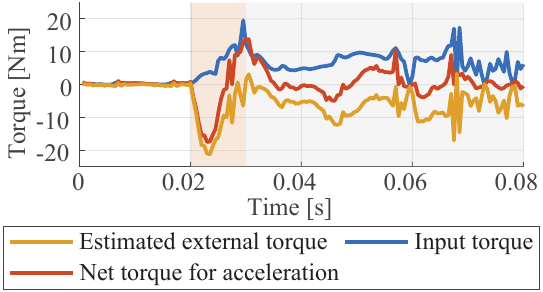}
    \caption{Experimental torque profiles at the ankle left motor during landing. Motor-side torque (without gear reduction) is shown.}
    \label{fig:external_torque_explanation}
\end{wrapfigure}

\textbf{External Torque at Landing.}  In hardware trials without transmission load regularization, the robot completed one front flip, but the ankle actuator’s sun gear fractured upon landing during the second trial, halting further experimentation. 
Fig.~\ref{fig:external_torque_explanation} shows the estimated external torque during landing, computed as the difference between input torque and net torque, with the latter derived from measured joint acceleration. As contact forces are not directly measurable on hardware, this provides an indirect estimate of the load on the actuator.
During the initial impact phase (10 ms after contact, 0.02–0.03 s in Fig.~\ref{fig:external_torque_explanation}), a substantial external load was transmitted to the actuator. Although the policy included contact impulse regularization, it failed to mitigate the external load during impact, resulting in gear fracture without transmission load regularization.

\begin{figure}[t]
  \centering
  \includegraphics[width=0.98\columnwidth]{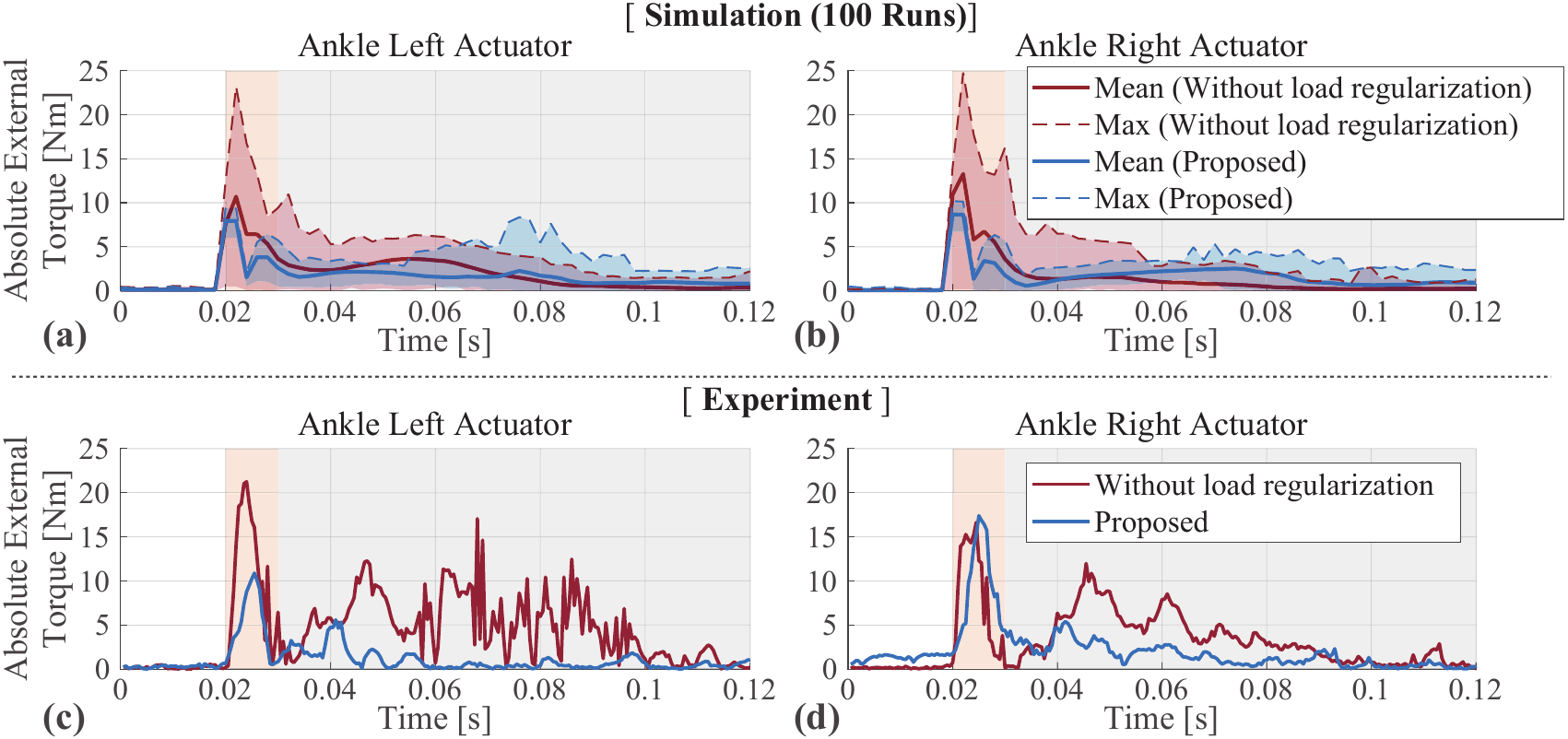}
  \caption{External torques on the ankle actuator’s sun gear during landing, for policies trained without vs.\ with transmission load regularization. (a), (b) are simulated; (c), (d) are estimated from hardware experiments. Time was aligned such that landing starts at 0.02 s in all cases.}
  \label{fig:external_torque_sim_and_real}
\vspace{-10pt}
\end{figure}

\textbf{Simulation and Hardware Evaluation.} 
We evaluate whether transmission load regularization reduces stress on vulnerable transmission components, specifically the sun gear of the ankle actuator, by comparing policies trained with and without this regularization. 
Fig.~\ref{fig:external_torque_sim_and_real} shows the estimated external torque at the ankle motor-side (i.e., on the sun gear), from simulation and hardware experiments. 
In the simulation results (Fig.~\ref{fig:external_torque_sim_and_real}a–b), solid lines show the mean external torque, and shaded regions indicate the min–max range across runs.

The policy without transmission load regularization exhibits a broader min–max range of external torque across simulation runs (Fig.~\ref{fig:external_torque_sim_and_real}a–b). This arises from flat-foot landings, where slight deviations in foot orientation cause the foot to skid or rotate unpredictably upon impact, resulting in irregular and asymmetric load transmission through the ankle. The expanded external torque range can impose excessive mechanical load on transmission components. In contrast, the policy trained with regularization yields lower and more consistent external torque trajectories. The learned strategy adjusts ankle orientation to initiate contact at a designated foot corner, followed by a rolling motion through pitch and roll after impact. This extends the contact duration and distributes impact forces over time. Although the mean external torque is comparable, Fig.~\ref{fig:external_torque_sim_and_real}a–b show that the regularized policy reduces variability and yields lower peak external torques. This improvement can reduce the likelihood of excessive peak stress on transmission components during high-impact landings. 

\begin{wrapfigure}{r}{0.4\textwidth}     
\centering
  \includegraphics[width=0.4\columnwidth]{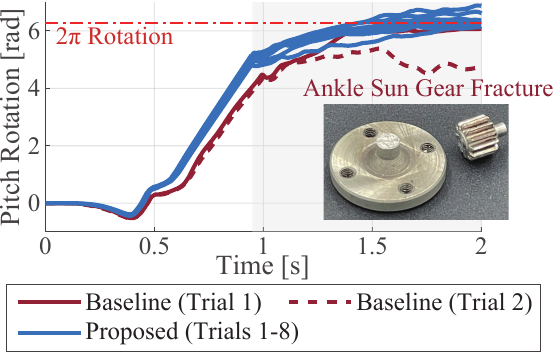}
  \caption{Pitch rotation across repeated hardware flip trials. The baseline (without load regularization) failed on the second trial due to sun gear fracture (inset photo); the regularized policy completed eight trials without failure.}
  \label{fig:all_trials}
\vspace{-10pt}
\end{wrapfigure}

The effect of transmission load regularization is further validated through hardware experiments. As shown in Fig.\ref{fig:external_torque_sim_and_real}c–d, the estimated peak external torque decreases from approximately 21 N$\cdot$m to 17 N$\cdot$m with regularization, and the peak occurs later in time, indicating more compliant or delayed impact absorption. This reduction in transmission stress translates to improved hardware robustness. Fig.~\ref{fig:all_trials} shows pitch rotation over repeated trials: the non-regularized policy caused sun gear fracture on the second attempt, whereas the regularized policy completed eight trials without mechanical failure. These results demonstrate that transmission load regularization not only reduces peak stress but also improves consistency across varied conditions, yielding substantial gains in durability for impact-rich maneuvers.

\section{Conclusion}
\label{sec:conclusion}
This paper presents a framework for learning and deploying impact-rich, rotation-intensive behaviors in legged robots using centroidal velocity–based rewards and actuator-aware sim-to-real techniques. The proposed centroidal angular velocity reward enables whole-body rotations beyond the limitations of conventional link-level rewards. Incorporating Motor Operating Region (MOR) modeling and transmission load regularization enhances the sim-to-real reliability of learned policies by respecting real-world actuator constraints and improving structural robustness. We validated the proposed framework by achieving the first hardware front flip on a one-leg hopper, a minimal yet highly challenging platform. This task, demanding large aerial rotations and high-magnitude impact absorption, served as a representative benchmark for testing the robustness of our approach.

\clearpage

\section*{Limitations}

The proposed framework has been evaluated primarily on a single hardware platform—a custom-designed one-leg hopper—and focused on a specific dynamic maneuver, namely the front flip. While this setting serves as an effective benchmark for testing impact-rich rotational motion, it does not fully represent the diversity of behaviors or morphologies found in more complex legged systems. 
Furthermore, our reward formulation and control strategies have not yet been validated on other motion types such as spins, twists, or sequential acrobatic actions.
Nonetheless, the underlying principles introduced in this work are not limited to the one-leg hopper scenario. These components are broadly applicable to other legged platforms and dynamic tasks, and we believe they offer promising building blocks for future applications in agile locomotion and acrobatic robotics.

\acknowledgments{This work was supported by the Technology Innovation Program(or Industrial Strategic Technology Development Program-Robot Industry Technology Development)(RS-2024-00427719, Dexterous and Agile Humanoid Robots for Industrial Applications) funded By the Ministry of  Trade Industry \& Energy(MOTIE, Korea)
}

\bibliography{refs}

\clearpage

\appendix
\section{Appendix}

\subsection{Reward Formulations} \label{app:reward formulation}
\begin{table}[h]
\centering
\small
\begin{threeparttable}
\caption{Reward terms with expressions and variable definitions.}
\label{tab:reward_table}
\begin{tabular}{p{0.6cm}p{7.3cm}p{4.6cm}}
\toprule
\textbf{Term} & \textbf{Expression} & \textbf{Key variables and description} \\
\midrule
\multicolumn{3}{l}{\textit{Motion Reward Terms}} \\
\midrule
$r_{\mathrm{CAV}}$ & 
\shortstack[l]{
$3\cdot
\begin{cases}
0, & 0\le\phi<\phi_{\text{jump}}\\
\max(\min(\boldsymbol{\alpha}^T \mathbf{w}_{\mathrm{com}}, 10), -0.1), &\phi_{\text{jump}}\le\phi<\phi_{\text{land}}\\
-0.5\,\min(\|\mathbf{L}_{\mathrm{com}}\|,\,2.5), & \phi_{\text{land}}\le\phi<2.0
\end{cases}$
} 
& 
\begin{tabular}[t]{@{}l@{}}
$\boldsymbol{\alpha}$: target flip axis \\
$\mathbf{w}_{\mathrm{com}}$: centroidal angular velocity \\
$\mathbf{L}_{\mathrm{com}}$: angular momentum \\
$\phi$: motion phase \\
\end{tabular}\\
$r_{\mathrm{lin}}$ & $0.10\exp(-\|\mathbf{v}_{xy}\|^2)$ & $\mathbf{v}_{xy}$: horizontal base velocity \\
$r_{\tau}$ & $0.10\exp(-5\!\times\!10^{-3}\|\mathbf{W}_{\tau}\boldsymbol\tau\|)$ & $\boldsymbol\tau$: joint torque\\
$r_{p}$ & $0.10\exp(-5\!\times\!10^{-2}\|\mathbf{W}_{p}(\theta - \theta_0)\|^2)$ & $\theta$: joint positions; $\theta_0$: reference pose \\
$r_{v}$ & $0.20\exp(-5\!\times\!10^{-3}\|\mathbf{W}_{v}\dot\theta\|^2)$ & $\dot\theta$: joint velocities \\
$r_{a}$ & $0.20\exp(-1\!\times\!10^{-3}\|\mathbf{W}_{a}(\dot\theta - \dot\theta_{\mathrm{prev}})\|^2)$ & $\dot\theta_{\mathrm{prev}}$: previous joint velocity \\
$r_{\mathrm{slip}}$ & $0.10\exp(-0.2\|\mathbf{v}_{\mathrm{foot}}\|^2)$ & $\mathbf{v}_{\mathrm{foot}}$: foot velocity (xy plane) \\
$r_{\mathrm{act}}$ & $0.20\exp(-0.18\|\Delta a\|^2)$ & $\Delta a = a_t - 2a_{t-1} + a_{t-2}$: second-order action difference \\
$r_{\mathrm{cs}}$ & $0.10\exp(-\|c_t - c_{t-1}\|^2)$ & $c_t$: contact state vector \\
$r_{\mathrm{ci}}$ & $0.20\exp(-100\|\lambda\|^2)$ & $\lambda$: contact impulse vector \\
\midrule
\multicolumn{3}{l}{\textit{Barrier Reward Terms (see Appendix~\ref{app:relaxed log barrier})}} \\
\midrule
$b_{\mathrm{pos}}$ & $\sum_{j} b(\theta_j;\underline{\theta}_j,\overline{\theta}_j,0.08)$ & $\theta_j$: joint positions with limits\\
$b_{\mathrm{vel}}$ & $\sum_{j} b(\dot\theta_j;\underline{\dot\theta}_j,\overline{\dot\theta}_j,2.0)$ & $\dot\theta_j$: joint velocities with limits\\
$b_{\mathrm{load}}$ & $b(\tau_{\mathrm{load}};-30,30,1)$ & $\tau_{\mathrm{load}}$: joint load from contact (see Sec.~\ref{subsec:simtorealtechniques}) \\
\bottomrule
\end{tabular}
\begin{tablenotes}
\scriptsize
\item \textbf{Note.} 
$\mathbf{W}_{\tau}$, $\mathbf{W}_p$, $\mathbf{W}_v$, $\mathbf{W}_a$ are diagonal weight matrices for the 3 actuated joints: knee, ankle pitch, and ankle roll. Specifically, $\mathbf{W}_{\tau} = \operatorname{diag}(0.5,\ 2.0,\ 2.0)$, $\mathbf{W}_{p} = \operatorname{diag}(0.5,\ 1.0,\ 1.0)$, $\mathbf{W}_{v} = \operatorname{diag}(0.5,\ 2.0,\ 2.0)$, $\mathbf{W}_{a} = \operatorname{diag}(0.5,\ 2.0,\ 2.0)$.
Joint limits used in the barrier rewards are
$\underline{\theta} = [-0.4,\ -0.873,\ -0.4]$, $\overline{\theta} = [2.15,\ 0.698,\ 0.4]$, $\underline{\dot\theta} = [-8,\ -12,\ -12]$, $\overline{\dot\theta} = [8,\ 12,\ 12]$.
\end{tablenotes}
\end{threeparttable}
\end{table}

Our total reward consists of two parts: \emph{motion rewards}, which guide the discovery and regularization of the motion, and \emph{barrier rewards}, which enforce safety and hardware limits.
The motion reward combines the centroidal angular velocity term from Sec.~\ref{subsec:centroidalvelocityreward} with a collection of regularization penalties that suppress excessive or unstable behaviors. 
The barrier reward uses a relaxed log barrier~\cite{relaxedLogBarrier2024} to sharply penalize violations of joint, velocity, and transmission limits.

The overall scalar rewards are defined as:
\[
r_{\mathrm{motion}}
=\bigl(1 + r_{\mathrm{CAV}}\bigr)\sum_{i}w_{i}\,r_{i},
\qquad
r_{\mathrm{barrier}}
=\sum_{j}w_{j}\,b_{j},
\]
where \(r_{\mathrm{CAV}}\) is the centroidal angular velocity term, \(r_{i}\) are the auxiliary motion‐regularization rewards with weights \(w_{i}\), and \(b_{j}\) are barrier terms defined via relaxed log barrier function \(b(\cdot)\) (Appendix \ref{app:relaxed log barrier}) with weights \(w_{j}\).

The key terms are summarized below:
\begin{itemize}
  \item \(r_{\mathrm{CAV}}\): centroidal angular velocity(CAV) reward (Sec.~\ref{subsec:centroidalvelocityreward}).  
  \item \(r_{\mathrm{lin}}\): horizontal linear velocity reward.
  \item \(r_{\tau}\): torque penalty to encourage energy-efficient motions.  
  \item \(r_{p},\,r_{v},\,r_{a}\): joint position, velocity, and acceleration regularizations for smooth trajectories.  
  \item \(r_{\mathrm{slip}}\): foot slip penalty to minimize contact sliding.  
  \item \(r_{\mathrm{act}}\): action smoothness penalty on consecutive commands.  
  \item \(r_{\mathrm{cs}}\): contact smoothness penalty to avoid abrupt contact mode changes.  
  \item \(r_{\mathrm{ci}}\): contact impulse penalty to discourage reliance on large simulated impulses. 
  
  \item \(b_{\mathrm{pos}}\): joint position barrier to enforce position limits.  
  \item \(b_{\mathrm{vel}}\): joint velocity barrier to enforce velocity limits.  
  \item \(b_{\mathrm{load}}\): transmission load barrier (Sec.~\ref{subsec:simtorealtechniques}).  
\end{itemize}

Detailed expressions for each reward term are provided in Table~\ref{tab:reward_table}.

\subsection{Relaxed Log Barrier Function} \label{app:relaxed log barrier}
The relaxed log barrier function is used to softly enforce inequality constraints while retaining differentiability near the boundary. This specific formulation is adapted from~\cite{relaxedLogBarrier2024}, which introduces a smooth approximation to the standard logarithmic barrier.

The relaxed log barrier term is defined for a scalar \(x\) with lower and upper limits \(\alpha_{\min},\alpha_{\max}\) and relaxation parameter \(\delta\) as
\[
b(x;\alpha_{\min},\alpha_{\max},\delta)
=\,f(x-\alpha_{\min}) + f(\alpha_{\max}-x),
\]
where
\[
f(z) =
\begin{cases}
\displaystyle
\log\delta-\frac{1}{2}\Bigl(\bigl(\tfrac{z-2\delta}{\delta}\bigr)^{2}-1\Bigr)\;,
& z < \delta,\\[8pt]
\log z, & z \ge \delta.
\end{cases}
\]

\subsection{Network Architecture and Training Details.}  \label{app:training}
The policy network consists of three fully connected hidden layers of size 256, 128, and 64, each followed by a LeakyReLU activation.
It is trained using Proximal Policy Optimization (PPO) with an asymmetric actor–critic formulation. The actor’s observation includes the gravity vector in the body frame, the body’s angular velocity from the IMU, the current angles and angular velocities of the left and right ankle motors, and an action-history buffer of the two most recent commands. To capture temporal context, we also include the joint position error and joint velocity history over the past 0.18 s as well as the sine and cosine of the phase variable.
The actor outputs target joint positions, which are passed to a low-level proportional-derivative (PD) controller to generate torque commands according to \(\tau = K_p (q^* - q) - K_d \dot{q}\),  
where \(q^*\) is the target joint position given by the policy, \(q\) is the current joint position, and \(\dot{q}\) is the joint velocity.  
This torque is then applied to the simulated or real robot.
The critic receives both the actor’s observation and privileged data, including the unactuated ankle’s pitch and roll joint positions and velocities, and the body’s linear velocity. Its value function is realized as a dual-headed MLP, where each head consists of three hidden layers with sizes 256, 128, and 64.  One head predicts the return associated with the primary motion reward, while the other predicts the return for the barrier penalty. We compute two advantage functions, one from each head, and form the clipped surrogate loss using their sum with equal weighting. This formulation yields more accurate return estimates and provides stronger guidance to the actor.
To bridge the information gap, we leverage a concurrent estimation framework. An auxiliary estimator network, implemented as a two-layer MLP with hidden sizes 256 and 128, ingests the actor’s observations and predicts the privileged states used by the critic. 
These estimated states are concatenated with the original observations and fed into the policy network, relieving the actor from modeling the full closed-loop dynamics from scratch.

\subsection{Auxiliary Methods for Sim-to-Real Transfer} \label{app:auxiliary sim2real}
In addition to our core sim-to-real techniques, we also adopt auxiliary strategies during training to further enhance transferability to the real hardware.

\textbf{Domain Randomization and Noise Injection.}  
We apply domain randomization and noise injection to improve robustness against modeling errors and hardware variability.  
At the start of each episode, we randomize key physical properties (mass, inertia, center of mass), environmental parameters (friction, restitution), actuator properties, and sensor noise.  
These perturbations expose the policy to diverse conditions and improve transfer stability.  
Detailed randomization and noise parameters are listed in Table~\ref{tab:randomization} and Table~\ref{tab:obs_noise}.

\begin{table}[h]
  \centering
  \caption{Domain randomization settings used during training.}
  \label{tab:randomization}
  \small
  \begin{tabular}{ll}
    \toprule
    \textbf{Parameter} & \textbf{Distribution / Range} \\
    \midrule
    Total mass & Uniform scaling in [0.85, 1.15] \\
    Link inertias & Uniform scaling in [0.85, 1.15] \\
    COM position & Uniform shift in [-3, 3]cm \\
    Joint PD gains & Uniform scaling in [0.9, 1.1] \\
    Joint friction & Uniform in [0, 0.7] Nm (knee), [0, 1.0] Nm (ankle) \\
    Ground friction coefficient & Uniform in [0.5, 0.9] \\
    Restitution coefficient & Uniform in [0, 0.2] \\
    \bottomrule
  \end{tabular}
\end{table}
\begin{table}[h]
  \centering
  \caption{Observation noise settings (uniform noise applied with specified scales).}
  \label{tab:obs_noise}
  \small
  \begin{tabular}{ll}
    \toprule
    \textbf{Observation Type} & \textbf{Noise Scale} \\
    \midrule
    Gravity vector (body frame)  & 0.07 \\
    Body angular velocity & 0.10 rad/s \\
    Joint position & 0.05 rad \\
    Joint velocity & 0.50 rad/s \\
    Joint position history & 0.05 rad \\
    Joint velocity history & 0.10 rad/s \\
    \bottomrule
  \end{tabular}
\end{table}

\textbf{Contact Mode and Contact Impulse Regularization.}  
To mitigate reliance on inaccurate impact modeling, we penalize excessive contact impulses during takeoff and landing.  
Simulated contact forces are governed by simplified restitution and friction models that often diverge from real-world behavior.  
As a result, policies that exploit large simulated impulses can fail on hardware, particularly when real-world discrepancies cause contact events to be mistimed or weakened.  
In particular, a policy may learn to rely on a single high-impulse contact occurring over a very short time window.
In reality, even a slight timing mismatch can cause that critical contact to be missed, leading to failure of the entire maneuver.  
To prevent this, we penalize both the magnitude of each contact impulse and abrupt changes across timesteps, reducing high-frequency spikes that are unlikely to be realizable on physical hardware.

\subsection{Validation of External Torque Estimation.}  
\vspace{-1.5mm} 
\begin{figure}[h]
\centering
\includegraphics[width=0.95\textwidth]{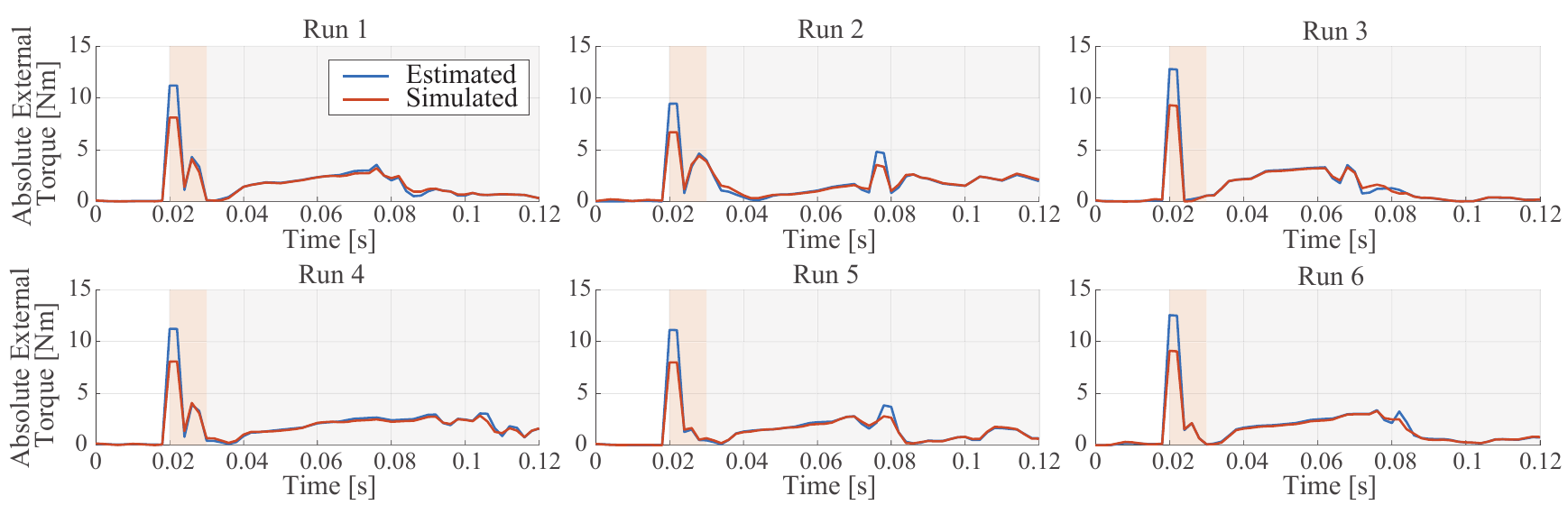}
\caption{Comparison of estimated and simulated external torque on the ankle actuator across six representative trials during impact.}
\label{fig:estimation_accuracy}
\end{figure}

We estimate the external torque as the difference between the input torque and the torque required to accelerate the effective inertia, computed as: $\boldsymbol{\tau}_{\mathrm{ext}}=\boldsymbol{\tau}_{\mathrm{input}} - \mathbf{I}_{\mathrm{eff}}~\boldsymbol{\dot\omega}_{\mathrm{rotor}} $.
Since the ankle actuator is connected to the foot through a closed-loop mechanism, we approximate the effective inertia as the sum of the rotor inertia and half of the foot link’s inertia : $\mathbf{I}_{\mathrm{eff}} = \mathbf{I}_{\mathrm{rotor}}+\mathbf{I}_{\mathrm{foot}}/2$ .

To evaluate the accuracy of this estimation method, we compared the estimated torques against ground-truth values computed from contact forces in simulation (see Sec.~\ref{subsec:simtorealtechniques}). 
Figure~\ref{fig:estimation_accuracy} illustrates six representative trials, overlaying the estimated and simulated torques. The shaded orange region indicates the initial impact phase (0.02–0.03 s), and the gray region marks the post-impact phase (0.03–0.12 s). Across these trials, the estimated torques closely match the simulated ground truth in both timing and magnitude.

To further quantify accuracy, we compute root-mean-square error (RMSE) within each phase, averaged across 100 trials and both ankle actuators. 
The resulting RMSE is 1.783 Nm during the initial impact phase and 0.297 Nm during the post-impact phase, indicating reasonable alignment between estimated and simulated torques throughout the contact event.

\subsection{Applicability to Other Platforms and Tasks.}  
To evaluate the generality of our framework, we extended its application beyond the front flip on a one-leg hopper to other tasks and a more complex robotic platform.

\begin{figure}[htb]
\centering
\begin{subfigure}{0.95\textwidth}
\centering
\includegraphics[width=\textwidth]{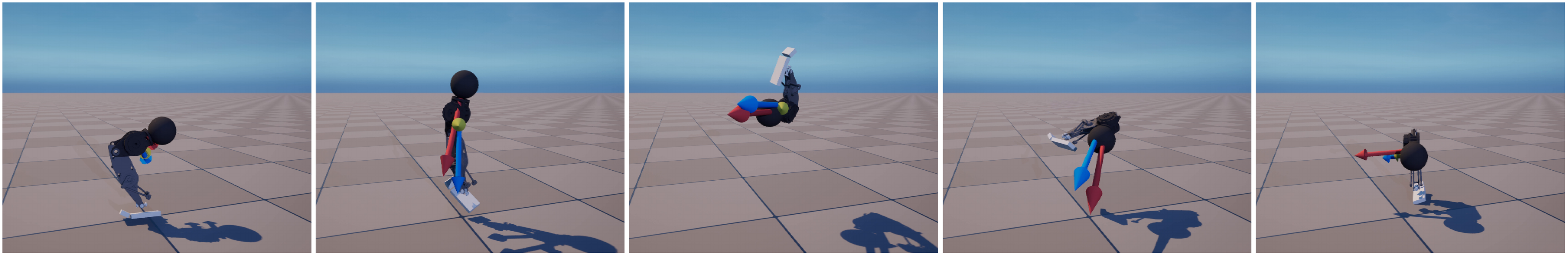}
\caption{Barrel roll}
\label{fig:hopper_barrel_roll}
\end{subfigure}
\begin{subfigure}{0.95\textwidth}
\centering
\includegraphics[width=\textwidth]{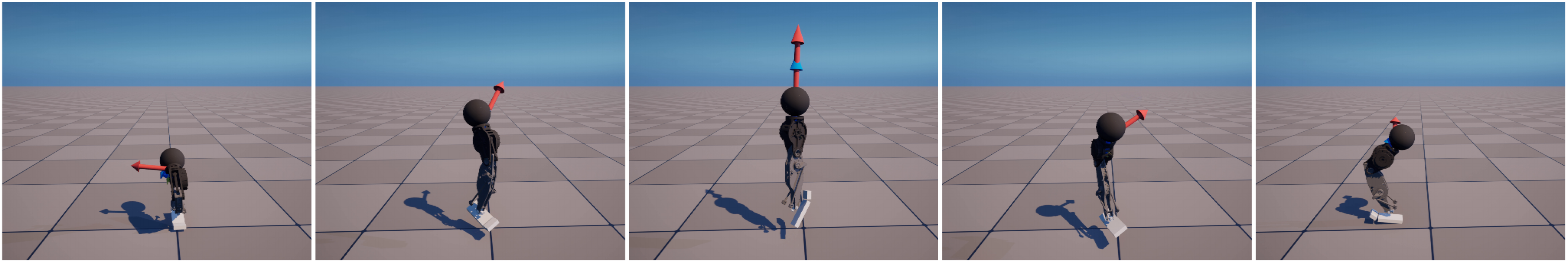}
\caption{Yaw-spin}
\label{fig:hopper_yaw_spin}
\end{subfigure}
\vspace{-0.5mm}
\caption{Additional maneuvers learned by the one-leg hopper using the same reward framework. }
\label{fig:hopper_other_motions}
\end{figure}

First, we tasked the one-leg hopper, now operating without its closed-loop ankle mechanism, with learning two additional aerial maneuvers: a yaw-spin and a barrel roll. As shown in Fig.~\ref{fig:hopper_other_motions}, both motions were successfully learned by merely adjusting the desired rotational axis vector in the reward function. This demonstrates the flexibility of our centroidal angular velocity reward, as no other changes to the reward formulation or learning pipeline were necessary.

\begin{figure}[htb]
\centering
\includegraphics[width=0.95\textwidth]{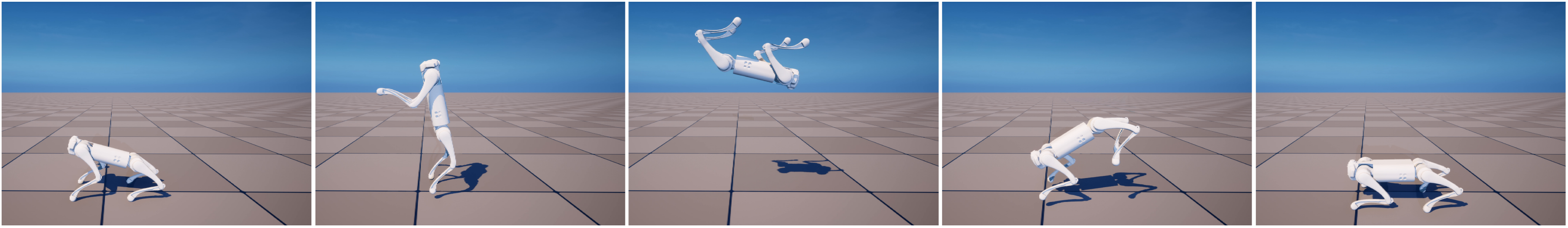}
\caption{A backflip trained on the Unitree Go1 quadruped.}
\label{fig:Go1_CAV_backflip}
\end{figure}
Furthermore, we applied our framework to a morphologically distinct platform, the Unitree Go1 quadruped, to train a backflip policy in simulation (Fig.~\ref{fig:Go1_CAV_backflip}).
The policy was successfully acquired using the identical centroidal angular velocity reward structure and actuator-aware constraints (MOR modeling) from the hopper experiments. 
This result indicates that our approach generalizes effectively across platforms with different morphologies and actuation structures.

These additional results confirm that our method is capable of scaling to a wide range of dynamic, impact-rich maneuvers across various legged systems.


\end{document}